\documentclass[sigconf,natbib=true]{acmart}

\AtBeginDocument{%
  \providecommand\BibTeX{{%
    \normalfont B\kern-0.5em{\scshape i\kern-0.25em b}\kern-0.8em\TeX}}}

\copyrightyear{2024}
\acmYear{2024}
\setcopyright{acmlicensed}\acmConference[WWW '24 Companion]{Companion Proceedings of the ACM Web Conference 2024}{May 13--17, 2024}{Singapore, Singapore}
\acmBooktitle{Companion Proceedings of the ACM Web Conference 2024 (WWW '24 Companion), May 13--17, 2024, Singapore, Singapore}
\acmDOI{10.1145/3589335.3651230}
\acmISBN{979-8-4007-0172-6/24/05}

\usepackage{todonotes}
\usepackage{amsmath}
\usepackage{amsfonts}
\usepackage{footnote}
\begin{document}

\title{Automating the Information Extraction from Semi-Structured Interview Transcripts}
% \todo{Cooler name?}


\author{Angelina Parfenova}
\affiliation{%
  \institution{Lucerne University of Applied Sciences and Arts}
  \streetaddress{1 Th{\o}rv{\"a}ld Circle}
  \city{Rotkreuz}
  \country{Switzerland}}
\email{angelina.parfenova@hslu.ch}

\renewcommand{\shortauthors}{Angelina Parfenova}

\begin{abstract}
This paper explores the development and application of an automated system designed to extract information from semi-structured interview transcripts. Given the labor-intensive nature of traditional qualitative analysis methods, such as coding, there exists a significant demand for tools that can facilitate the analysis process. Our research investigates various topic modeling techniques and concludes that the best model for analyzing interview texts is a combination of BERT embeddings and HDBSCAN clustering. We present a user-friendly software prototype that enables researchers, including those without programming skills, to efficiently process and visualize the thematic structure of interview data. This tool not only facilitates the initial stages of qualitative analysis but also offers insights into the interconnectedness of topics revealed, thereby enhancing the depth of qualitative analysis.

\end{abstract}




% %% A "teaser" image appears between the author and affiliation
% %% information and the body of the document, and typically spans the


\maketitle

\section{Introduction}
Qualitative methods such as interviews or focus groups with customers are an integral part of the research arsenal in a number of fields: marketing, social science, and medical studies \cite{Avjyan2005, Leeson2019}. This approach differs significantly from quantitative techniques in its ability to draw on individual experiences and delve deeper into the issue under study. However, unlike the results of quantitative surveys, in interviews, there is no ready-made information, no statistics, and no clear answers to the questions posed. The researcher unwittingly faces the problem of interpretational objectivity, and the question arises as to how to tackle it.

\begin{figure}
    \centering
    \includegraphics[width=\linewidth]{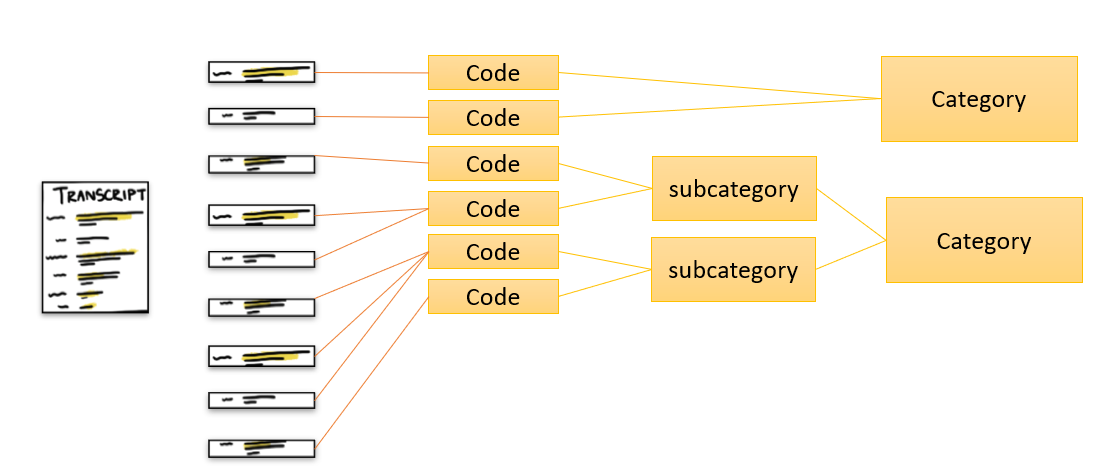}
    \caption{The coding process visualized}
    \label{fig:coding_process1}
\end{figure}

The general analysis of collected data in interviews mainly uses open coding technology (Fig \ref{fig:coding_process1}), which involves repeatedly reading the text to identify "codes" that are in essence important thoughts, ideas, attitudes, and subjects. Further, the axial coding procedure is applied, where the relationships between the codes and their aggregation into higher-level categories are found \cite{Saldana2016}. Several other coding methods are present and all of them involve independent work with the text, consisting of re-reading and finding the key thoughts of the informant in a large number of documents. This process often takes several weeks \cite{Alshenqeeti2014}. Hence, it can be seen that this procedure requires a lot of human effort to process the text by oneself. So the research issue of implementing automatization of the whole process or pre-processing of the text corpus to facilitate the subsequent analysis appears. 

That’s why the  main goal of current research is to automate the analysis of qualitative research results and elaborate the appropriate software that will help organizations and researchers dealing with large clients’ text corpora. To begin with, it is necessary to consider the existing solutions on the market and describe how the future service will differ.

\section{Current coding practices}

Each statement or significant segment of dialogue within an interview is assigned a 'code' that summarizes its main idea. Codes are not just words, phrases, sentences, or thoughts but represent a unit of meaning that encapsulates key aspects of the data \cite{miles1994qualitative}. Once coded, these segments are then organized into broader categories that reflect the underlying patterns and relationships within the dataset \cite{glaser2017discovery}. In practice categories and codes consist of one or two words to encapsulate the main meaning of a citation. However if the main thought of a sentence can be described only in a phrase, that is also allowed.

Describing the process in simpler terms, first, we summarize the main idea of each citation in the interview\cite{RyanBernard2003}. Then, we start grouping them into bigger categories. This means looking at all the little ideas we've found and seeing how they fit together into larger themes. We ask questions like, "Do these codes share something in common?" or "Are they talking about the same bigger idea?" This helps us organize our findings better. 

These categories serve as the pillars for constructing a conceptual framework \cite{Lochmiller2021ThematicAnalysis}, which researchers often visualize in the form of a graph. Such a graph, akin to a mind map, interlinks individual responses, highlighting the associations and hierarchies amongst different thematic codes. This visualization assists in better understanding the collective narrative of the participants. Thus, the coding process is a critical interpretive phase in qualitative research, helping in developing conclusions and theoretical insights.

\section{Overview of methods and tools}
\subsection{Existing softwares}
\paragraph{Text coding environments} In general, researchers use software to facilitate the coding of interviews as follows: the text is conveniently placed on the screen with the possibility of highlighting parts of the sentence with a color marker and tagging them with codes, the number of which then counts itself \cite{Atlas.ti, MAXQDA}. These programs also make space for drawing diagrams with codes and connecting them with arrows and written relationships. Thus, the software does not replace the analysis process as such but simply puts the researcher in a comfortable environment for the same workflow.
\paragraph{Text analysis software} When it comes to programs aimed at replacing some of the work some softwares replace part of the work by giving analytical tools such as word statistics and word cloud \cite{Dedoose}. It is worth noting that all of these programs are paid, so not all researchers are inclined to use them. For example, in one of the works already described, coding was done in MS Word and Excel \cite{Leeson2019}. Some programs focus on text processing in general, with clustering and collocation search capabilities \cite{AntConc}. However, the format of interview analysis is very specific, as it requires the building of models based on a set of answers to one question from several informants. Therefore, it is difficult to use such general-purpose software for the analysis of transcripts.

Thus, the purpose of this work is to develop a method for analyzing transcripts of qualitative research results, as well as to write user-friendly software that can be used by researchers who do not know the skills of programming. 

\section{Research design}

In this paper, several methods will be applied and compared with each other, and finally, the most appropriate method for analyzing qualitative research transcripts will be chosen. To achieve the goal, it is necessary to perform the following tasks:

\begin{itemize}
    \item Compare different methods of topic modeling and choose the best one for semi-structured interview transcripts
    \item Create a framework for visualizing selected keywords from transcripts
    \item Write a frontend understandable to researchers
\end{itemize}

\subsection{Analysis of interview transcript data and selection of the best topic model}

\paragraph{Specifics of interview texts: preprocessing}

This paper will compare various methods to find the most suitable one for analyzing interview transcripts. Typically, topic modeling \cite{Blei2012} is used for either large texts with multiple topics or very short texts like tweets or reviews. Semi-structured interviews present a unique challenge as they include characteristics of both: they are short like tweets but can contain rich narratives similar to longer documents, making standard topic modeling approaches less effective.

Before constructing topic models, the necessary preprocessing of the natural text is done. First, the sentences were tokenized and lemmatized. Stop words, which are included by default were also removed. However, in addition to this, a complementary set of stop words was created, which was compiled independently after building the frequency tables of the tokens. The point is that many undesirable words occurring in interviews could be merged into one topic (this happened in one of the LDA models). Such words as ” probably, it turns out, in general, supposedly, like” were removed, as well as some verbs that refer to the process of reflection ”think, suppose”.

\paragraph{Specifics of interview texts: document structure}

First, it is possible to combine the answers to the questions in one large document and build topic models based on the combined answers to one question. Or it is possible to use all of the transcripts and build one large model. In this work, both methods were tested on one set of interviews and it was concluded that the topics obtained by answering one question gave unsatisfactory codes (denoted as W).

\begin{table}[H]
\centering
\caption{Partial topic modeling results for one interview question}
\begin{tabular}{l|l|l|l|l|l}
\hline
\textbf{Topic} & \textbf{W 0} & \textbf{W 1} & \textbf{W 2} & \textbf{W 3} & \textbf{W 4} \\ \hline
1 & work            & to work         & to study        & career          & electric     \\
2 & good            & knowledge       & to study        & saying          & succeed         \\
3 & work            & to work         & educate       & do              & receive         \\
4 & to work         & do              & work            & favorite        & business        \\ \hline
\end{tabular}
\label{tab:topics}
\end{table}

We can immediately notice in Table \ref{tab:topics} that all the received topics are very similar. This could have been foreseen since more often than not one question is about one specific topic. Therefore, it was decided to abandon this method of compiling documents right away. Thus, future model development was conducted on the aggregate of all documents. 

% \begin{figure*}
%     \centering
%     \includegraphics[width=\linewidth]{performance.png}
%     \caption{Enter Caption}
%     \label{fig:enter-label}
% \end{figure*}

\section{Experiments}

\paragraph{LDA} The initial approach utilized the standard LDA package from Gensim to establish a baseline for topic modeling of interview transcripts. The best-performing LDA model achieved moderate quality metrics but faced challenges in clearly distinguishing differentiated topics.

\paragraph{LDA+BERT} This experiment combined LDA with BERT embeddings to enhance contextual understanding of topics. Initial language for interview texts is Russian, which is why the Russian language BERT model from Deep Pavlov and an autoencoder for dimensionality reduction were used. For the English version of interviews, the BERT base uncased model was used. Clustering was done using K-means, but the model struggled with topic clarity and overall quality.

\paragraph{Top2Vec} The Top2Vec method was implemented using the same BERT model to get embeddings. UMAP was employed for dimensionality reduction, followed by clustering with HDBSCAN. This approach yielded more interpretable topics. Consequently, the BERT+UMAP+HDBSCAN algorithm was selected as the most suitable for semi-structured interviews, despite its longer processing time.

\begin{figure}
    \centering
    \includegraphics[width=0.8\linewidth]{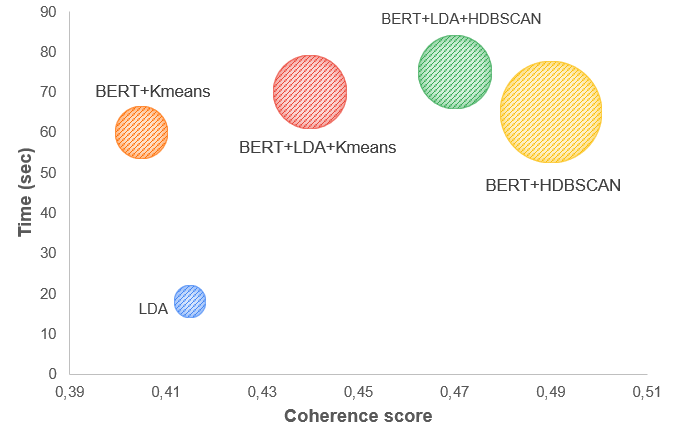}
    \caption{Comparison of model performance (Size of the bubble is Topic Diversity)}
    \label{fig:metrics}
\end{figure}

Across various sets of interviews, the BERT+HDBSCAN model consistently showed high topic diversity and better interpretability. However, for the prototype, model tuning was omitted in the frontend due to efficiency and time considerations. The model comparison for 10 topics and 2 sets of interviews is demonstrated in Table\ref{tab:model_comparison_consolidated} and Figure\ref{fig:metrics}. 

\begin{table*}[htbp]
\centering
\caption{Model Comparison for Two Sets of Interviews}
\begin{tabular}{l|cc|cc|cc|cc|cc}
\hline
 & \multicolumn{2}{c|}{\textbf{LDA}} & \multicolumn{2}{c|}{\textbf{BERT+LDA+Kmeans}} & \multicolumn{2}{c|}{\textbf{BERT+HDBSCAN}} & \multicolumn{2}{c|}{\textbf{BERT+LDA+HDBSCAN}} & \multicolumn{2}{c}{\textbf{BERT+Kmeans}} \\
\textbf{Metrics} & 1st & 2nd & 1st & 2nd & 1st & 2nd & 1st & 2nd & 1st & 2nd \\
\hline
C\_v            & 0.456 & 0.413 & 0.487 & 0.435 & 0.512 & 0.478 & 0.501 & 0.465 & 0.435 & 0.398 \\
Umass          & -12.354 & -11.346 & -3.29 & -3.15 & -3.56 & -4.18 & -6.763 & -5.476 & -2.891 & -4.021 \\
NPMI           & -0.289 & -0.453 & -0.034 & -0.023 & -0.042 & -0.011 & -0.112 & -0.098 & -0.056 & -0.021 \\
UCI            & -8.984 & -10.354 & -2.135 & -2.817 & -0.698 & -0.453 & -4.120 & -2.867 & -2.409 & -1.309 \\
Topic diversity & 0.679 & 0.690 & 0.890 & 0.789 & 0.950 & 0.984 & 0.835 & 0.905 & 0.756 & 0.740 \\
Silhouette     & NA & NA & 0.389 & 0.209 & NA & NA & NA & NA & 0.067 & 0.056 \\
DBCV           & NA & NA & NA & NA & 0.516 & 0.602 & 0.679 & 0.714 & NA & NA \\
\hline
\end{tabular}
\footnotesize{

Note: C\_v measures topic coherence; Umass and UCI are coherence scores; NPMI is normalized pointwise mutual information; Topic diversity indicates the uniqueness of topics; Silhouette measures cluster separation; DBCV is density-based clustering validation.}
\label{tab:model_comparison_consolidated}
\end{table*}

\section{The prototype}

The prototype was based on completed objectives. Firstly, the researcher uploads a
document with all the interviews to the website. Then he has a choice to either save or dismiss the Interviewer's phrases. This heavily depends on the researcher's perspective and type of interview. Then he has to press the lemmatize button and preprocess. Shortly after that, the program will give him the most frequently used words in the text, which he can load into the file ”additional stop words” if he wishes and add to the website again\footnote{\href{https://github.com/Likich/TM_graph}{https://github.com/Likich/TM\_graph}}.

\begin{figure}
    \centering
    \includegraphics[width=0.95\linewidth]{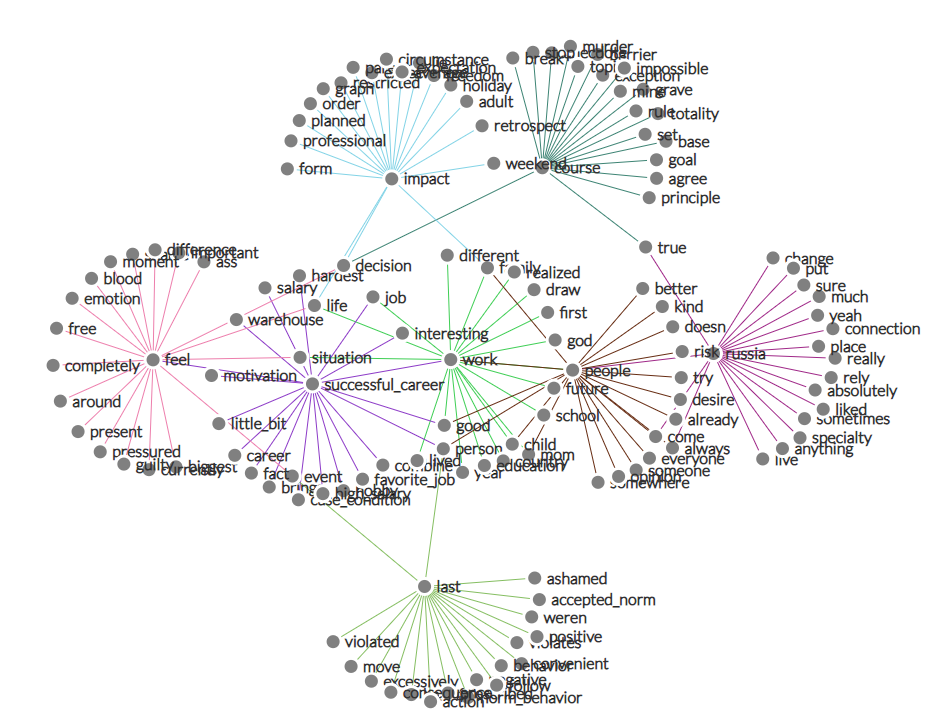}
    \caption{Output of the model on the set of interviews \cite{parfenova2023regulatory}}
    \label{fig:enter-label}
\end{figure}

Next, he can choose several suggested methods for analyzing the interview. LDA is the classic method and is fast, while BERT gives higher-quality results but takes more time. The user can also choose how many topics he would like to see in the result. Next, the researcher will be provided with an interactive graph of connections (Figure \ref{fig:enter-label}). At the moment it is only available in English and Russian languages, but in the future, there will be support for more languages.

Having the set of Topics: $T = \{t_1, t_2, ..., t_n\}$, the graph is constructed as follows: each vertex of the graph is a keyword that belongs to the topic $t_i$: $K_i = \{k_1, k_2, ..., k_m\}$. For each keyword in the topic we have its weight that demonstrates the importance of a word in describing the topic $t_i$: $W_i = \{w_1, w_2, ..., w_m\}$. 

Thus, the graph is constructed as follows:

\begin{equation}
    V = \bigcup_{i=1}^{n} \{v_j | k_j \in K_i\}
\end{equation}
\begin{equation}
    E = \{(v_j, v_k) | \text{keywords } k_j \text{ and } k_k \text{ co-occur within topics in } T\}
\end{equation}
\begin{equation}
    G = (V, E)
\end{equation}

In the middle is the keyword with the highest weight. The question arises as to whether the word with the highest weight characterizes the entire topic. The answer is no, but in practice, the word with the most weight in the topic is less likely to occur in other topics and is less likely to affect the quality of visualization. Moreover, using for example three keywords will not promise that they play a key role in determining the semantics of the topic, and visualization will be even more difficult.

\begin{equation}
    \text{For each } t_i, \text{ find } k_{max} \text{ where } k_{max} = \arg\max_{k \in K_i} W_i(k)
\end{equation}
\begin{equation}
    \text{Central vertex for } t_i: v_{max} \leftrightarrow k_{max}
\end{equation}

If a researcher is interested in knowing what citations mentioned this exact word, he can double-click the vertex of the network and citations will be shown. This phase is important to facilitate the interpretation of topics. 

According to this visualization, you can also see which topics are linked by co-occurrence and which are not. For example, in a series of interviews about social expectations, it was seen how the central keyword in the family topic, ”parent,” was related to ”education” as the central word in the education and work topic, and to ”shame,” the central word in the emotion topic. In this way, the researcher can understand which codes affect which, and which tops are completely disconnected and represent a separate topic. Based on these results, we can already hypothesize about the influence of indicators on others and prepare the methodology for the quantitative phase of the study.

\section{Empirical perspective}

Automating the coding process in qualitative research can be an advantage for various fields such as market research, customer feedback analysis, and clinical data analysis. These areas often deal with vast amounts of unstructured data like interviews, focus groups, and open-ended survey responses, where traditional manual coding is time-consuming and subject to human biases.

In market research automated coding can analyze customer interviews and group discussions faster, helping businesses find new trends and customer preferences more quickly. As for customer feedback analysis, automated coding can process customer feedback from various channels (social media, customer surveys, etc.) in real-time, enabling companies to respond promptly to customer needs and complaints. Analyzing qualitative feedback at scale allows for more personalized marketing strategies based on nuanced customer preferences and experiences. Creating a concept network from qualitative feedback can help in creating detailed customer journey maps, identifying pain points, and enhancing customer experience.

In healthcare, automated coding of patient interviews and feedback can provide insights into patient experiences, leading to improved care and treatment strategies reducing the time for data analysis, and accelerating research outcomes. Analysis of patient narratives and feedback can reveal insights into the efficacy of treatments and patient feedback, aiding in the improvement of therapeutic approaches.

\section{Conclusion}
The purpose of this study was to automate the analysis of semi-structured interviews, which is currently very time-consuming for individual researchers. It was also intended to write an application that would allow qualitative researchers with no knowledge of programming skills to use automatic text-processing methods. To achieve the goal this paper considered methods of coding texts, and considered the method of topic modeling, which identifies topics from the text with their keywords. Several methods of topic modeling were compared according to objective metrics and the most suitable one for semi-structured interview texts was selected.

The results showed that the best method is to convert the text into embeddings using BERT, then reduce the dimensionality of the resulting vectors using UMAP and clustering with HDBSCAN, followed by an algorithm to reduce the number of topics. For a basic assessment of model quality, it was decided to use the Topic Diversity metric, which will be important in constructing a pleasing and distinguishable visualization. After examining the front end, it became clear that with a high score on this metric, the graph is not cluttered with a large number of connections.

The theoretical significance of this work consists of testing more advanced and costly methods of interview analysis, whereas previous works mainly took the basic LDA, which proved to be the worst compared to methods using transformer architecture. The main practical contribution is that researchers in qualitative studies now will have access to automatic analysis of their work, or a convenient basis for subsequent analysis. It cannot be argued that such work completely replaces the individual researcher, who, firstly, is more familiar with the topic, and secondly, can analyze the truthfulness of the answers. However, automated analysis frees the researcher from his or her subjectivity and can help avoid judgmental attitudes.

\bibliographystyle{ACM-Reference-Format}
\bibliography{main}

\appendix

\end{document}